\documentclass[10pt,twocolumn,letterpaper]{article}

\usepackage[pagenumbers]{cvpr} 

\usepackage{algorithm}
\usepackage{algorithmic}
\usepackage{pifont}
\usepackage{multirow}
\definecolor{cvprblue}{rgb}{0.21,0.49,0.74}
\usepackage[pagebackref,breaklinks,colorlinks,allcolors=cvprblue]{hyperref}

\def\paperID{41436} 
\def\confName{CVPR}
\def\confYear{2026}

\title{FOZO: Forward-Only Zeroth-Order Prompt Optimization\\ for Test-Time Adaptation}

\author{Xingyu Wang \quad Tao Wang\thanks{Corresponding author.} \\
College of Computer Science, Sichuan University\\
{\tt\small wangxingyu0909@outlook.com \quad twangnh@gmail.com}
}

\begin{document}
\maketitle
\begin{abstract}

Test-Time Adaptation (TTA) is essential for enabling deep learning models to handle real-world data distribution shifts. However, current approaches face significant limitations: backpropagation-based methods are not suitable for low-end deployment devices, due to their high computation and memory requirements, as well as their tendency to modify model weights during adaptation; while traditional backpropagation-free techniques exhibit constrained adaptation capabilities.
In this work, we propose Forward-Only Zeroth-Order Optimization (FOZO), a novel and practical backpropagation-free paradigm for TTA. FOZO leverages a memory-efficient zeroth-order prompt optimization, which is led by objectives optimizing both intermediate feature statistics and prediction entropy.
To ensure efficient and stable adaptation over the out-of-distribution data stream, we introduce a dynamically decaying perturbation scale during zeroth-order gradient estimation and theoretically prove its convergence under the TTA data stream assumption.  Extensive continual adaptation experiments on ImageNet-C, ImageNet-R, and ImageNet-Sketch demonstrate FOZO's superior performance, achieving 59.52\% Top-1 accuracy on ImageNet-C (5K, level 5) and outperforming main gradient-based methods and SOTA forward-only FOA (58.13\%).  Furthermore, FOZO exhibits strong generalization on quantized (INT8) models. These findings demonstrate that FOZO is a highly competitive solution for TTA deployment in resource-limited scenarios. Code: https://github.com/eVI-group-SCU/FOZO
\end{abstract}    
\section{Introduction}
\label{sec:intro}
\begin{figure}[t]
  \centering
  \includegraphics[width=1.0\linewidth]{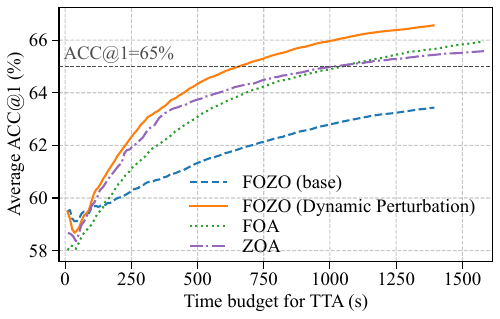}
  \caption{Convergence Curves of Forward-Only Test-Time Adaptation Algorithms. Average ACC@1 (\%) of various Test-Time Adaptation (TTA) methods on ImageNet-C (level 5) versus adaptation time (s). Per original settings, FOA and ZOA use 28 forward propagation (FP), while our FOZO method employs 26 FP. FOZO (Dynamic) consistently surpasses FOZO (Base) in accuracy, demonstrating the effectiveness of our dynamic perturbation strategy. Furthermore, FOZO achieves superior performance and faster convergence than FOA and ZOA. Notably, FOZO reaches 65\% ACC@1 in only 66\% of the runtime required by FOA and ZOA, respectively.}
  \label{fig:Comparison of Convergence Rates}
\end{figure}
Deep learning has revolutionized computer vision \cite{KrizhevskySH12AlexNet,GirshickDDM14RCNN}, achieving state-of-the-art performance in tasks. However, models often encounter data distribution shifts \cite{PanY10Transfer} between training and test environments in real-world deployment. To address this, Test-Time Adaptation (TTA) \cite{SunWLMEH20TTT,LiWS0H17AdaBN} has emerged as a promising paradigm, dynamically adjusting pre-trained models with unlabeled test data to improve generalization on unseen target domains.

Existing TTA methods have made significant progress \cite{NiuMCWZ24FOA, lee2024DeYO}. Notable approaches include gradient-free adaptation and gradient-based adaptation. The former typically works by adjusting the normalization of feature representations \cite{LiWS0H17AdaBN} or model outputs \cite{BoudiafMAB22LAME}. The methods following this approach maintain the original model weights and thus reduce the risk of forgetting the source domain. However, since they do not explicitly construct adaptation objectives based on the model feedback from the given test samples, their learning capacity is limited \cite{BurnsS21Limitations, ZhaoLAL23Pitfalls}, leading to suboptimal adaptation performance. Unlike the gradient-free approach, the gradient-based approach usually applies explicit optimization objectives \cite{SunWLMEH20TTT} to update the model weights, such as entropy minimization \cite{WangSLOD21TENT}. Several methods follow this approach and have achieved SOTA performance \cite{NiuW0CZZT22EATA,Niu00WCZT23SAR,lee2024DeYO}. Nevertheless, they still face challenges in practical deployment scenarios. These scenarios typically offer limited computation resources and memory to conduct the adaptation, or require hard-coded or quantized model weights that are difficult to update \cite{gholami2022survey}.

Recent research has explored methods that update learnable parameters solely through a forward pass, aiming to address challenges in existing TTA algorithms and develop more practical methods. The core advantage of this strategy is its reliance solely on the model's forward pass, obviating backpropagation and significantly reducing computational and memory overheads. An example is FOA\cite{NiuMCWZ24FOA}, which injects learnable input prompts and updates them with covariance matrix adaptation evolution strategy (CMA-ES) \cite{hansen2001completely}. However, despite the small number of learnable prompts, their high dimensionality still poses challenges for accurate update estimation with CMA-ES \cite{hansen2016cma}, leading to slow convergence during adaptation and suboptimal performance. ZOA \cite{ZOA} updates normalization layer parameters through zeroth-order optimization. But this method directly modifies internal model components, limiting its applicability in scenarios where model parameters are immutable \cite{prompttuning}.

Inspired by the zeroth-order gradient estimation technique \cite{J.C.92ZOSGD}, we propose a novel Forward-Only Zeroth-order Optimization (FOZO) paradigm for test-time adaptation. Instead of relying on CMA-ES to update prompts, FOZO relies on the model's forward pass to estimate prompt gradients, while avoiding the expensive backward propagation. Nevertheless, applying the zeroth-order forward gradient estimation in TTA scenarios faces severe optimization challenge: the out-of-distribution data stream leads to unreliable gradient estimates, hindering robust adaptation.

To address this problem, we introduce a dynamic perturbation scheme that adaptively adjusts perturbation scale within the zeroth-order gradient approximation. This helps the model quickly adapt to new domains, escaping initial suboptimal solutions while achieving stable convergence after initial adaptation. We theoretically show that the proposed zeroth-order prompt optimization method converges with the dynamic perturbation scheme (Theorem 1 and 2). Furthermore, to enable adaptation in unsupervised optimization environment, we designed a refined unsupervised loss function, including the model's prediction entropy and the activation statistical differences between source and target domain samples, which is computed from both the shallow and deep layers of the model. Extensive experimental results demonstrate that FOZO achieves superior performance compared to prior gradient-based methods while using only forward passes and maintaining low memory consumption. It also surpasses the SOTA forward-only adaptation method \cite{NiuMCWZ24FOA} with faster convergence. Notably, we further verify the effectiveness of FOZO under the practical deployment settings where models are quantized or test data arrives continually without knowing the domain boundary.

In summary, our contributions are:

\begin{itemize}
    \item We introduce a zeroth-order prompt optimization approach, based on which we build a novel forward-only test-time adaptation method that avoids updating parameters and achieves superior efficiency.
    \item To address the optimization challenge of zeroth-order test-time adaptation caused by distribution shifts in TTA data stream, we proposing a dynamic perturbation scheme to assist optimization.
    \item We theoretically prove the convergence properties of the proposed method, based on the classical stochastic perturbation approximation \cite{spsa} and local r-effective rank assumption \cite{MalladiGNDL0A23MeZO}.
    \item To validate the proposed method, we establish various test-time adaptation settings, and conduct thorough ablation and analysis. FOZO demonstrates superior performance to prior approaches and shows strong adaptation capability with the quantized models and practical continual TTA scenario.
\end{itemize}

\section{Related Work}
\subsection{Test-Time Adaptation}
The pioneering works of gradient-based TTA are TTT \cite{SunWLMEH20TTT} and OSHOT \cite{DInnocenteBBCT20OSHOT}, which implicitly align features from training and test domains by introducing self-supervised tasks during both phases. The strategy adopted and extended by subsequent methods \cite{ZhangNF20,HansenJSAAEPW21,YazdanpanahM22,wang2025test}. However, TTT requires modifying the model's training process, which has led subsequent works to explore fully test-time adaptation (TTA). A landmark is TENT \cite{WangSLOD21TENT}, it updates batch-norm parameters by minimizing output entropy, inspiring numerous extensions focused on stability \cite{Niu00WCZT23SAR}, additional self-supervised signals \cite{Dian22Contrastive, Kingetsu0OYN22,MirzaJ0KPB23ActMAD}, and anti-forgetting regularization \cite{WangFGD22CoTTA,NiuW0CZZT22EATA}.

Despite significant progress, their fundamental reliance on backpropagation limits deployment in resource-constrained edge devices or black-box models lacking gradient support. This fundamental limitation has driven the development of Forward-only TTA methods based entirely on forward propagation, which has garnered attention due to its high computational efficiency and ease of deployment, typically avoiding iterative optimization of core model parameters. Early works like AdaBN \cite{LiWS0H17AdaBN} and PredBN \cite{Nado20PredBN} primarily focused on BN layer calibration by replacing statistics. More sophisticated strategies include T3A \cite{IwasawaM21T3A}, which adapts by adjusting only the classifier head via online-updated ‘pseudo-prototypes’, and LAME \cite{BoudiafMAB22LAME}, which directly corrects output logits online through Laplace Approximated Maximum-a-Posteriori estimation without modifying model parameters. Recently, FOA \cite{NiuMCWZ24FOA} proposed a paradigm for optimizing model parameters without backpropagation, by employing an evolutionary algorithm \cite{hansen2001completely} to update learnable visual prompts at the ViT input layer.
\begin{algorithm}[tb]
\caption{Forward-Only Zeroth-Order Optimization for Test-Time Adaptation}
\label{alg:fozo_tta}
\textbf{Input}: Test batches $\{\mathcal{B}_t\}_{t=1}^T$, learning rate schedule $\eta_t$, initial perturbation scale $\epsilon_0$, minimal perturbation scale $\epsilon_{\text{min}}$, number of prompts $p$, number of SPSA samples $n$ \\
\textbf{Output}: The predictions $\{\mathcal{Y}_t\}_{t=1}^T$
\begin{algorithmic}[1]
\STATE Initialize Learnable prompt $\mathbf{P} = \{\mathbf{p}^k \in \mathbb{R}^d | k \in \mathbb{N}, 1 \leq k \leq p\}$, $\epsilon_t$ = $\epsilon_0$
\FOR{$t = 1, \ldots, T$}
    \STATE $\text{seeds, projected\_grads} \leftarrow []$
    \STATE Adjust the perturbation scale $\epsilon_t$ by Eqn. \ref{eq:dynamically adjust}
    \FOR{$j = 1, \ldots, n$}
        \STATE Generate $\mathbf{Z} \sim \mathcal{N}(0, I_d)$ using random seed $s$
        \STATE Compute $\mathbf{P}_{+} = \mathbf{P} + \epsilon_t \mathbf{Z}$ and $\mathbf{P}_{-} = \mathbf{P} - \epsilon_t \mathbf{Z}$
        \STATE Propagate $\mathcal{B}_t$ with $\mathbf{P}_{\pm}$ to obtain $\{\mathbf{e}_N^0\}_\pm$
        \STATE Predict $\{\hat{\mathcal{Y}}^j_t\}_\pm$ with $\{\mathbf{e}_N^0\}_\pm$
        \STATE Calculate $\ell_+ = \mathcal{L}(\mathbf{P}_+; \mathcal{B}_t)$, $\quad \ell_- = \mathcal{L}(\mathbf{P}_-; \mathcal{B}_t)$
        \STATE $\text{projected\_grad} = (\ell_{+} - \ell_{-}) / (2\epsilon_t)$
        \STATE $\text{projected\_grads}[j]\leftarrow\text{projected\_grad}$
        \STATE $\text{seeds}[j]\leftarrow s$
    \ENDFOR
    \FOR{$j = 1, \ldots, n$}
        \STATE Regenerate $\mathbf{Z}_j$  using $\text{seeds}[j]$
        \STATE Update $\mathbf{P}\leftarrow\mathbf{P} - (\eta_t/n) \cdot\text{projected\_grads}[j]\cdot\mathbf{Z}$
    \ENDFOR
    \STATE Select best  $\hat{\mathcal{Y}}_t$ from $\{\hat{\mathcal{Y}}^j_t\}_\pm$
\ENDFOR
\end{algorithmic}
\end{algorithm}
\subsection{Zeroth-Order Optimization}
Zeroth-order optimization (ZO) methods do not rely on explicit gradient, which makes them particularly useful when gradients are difficult, unavailable, or prohibitively expensive to compute. For instance, the classical zeroth-order stochastic gradient descent (ZO-SGD) \cite{J.C.92ZOSGD} estimates gradients by performing random perturbations in the parameter space and comparing the function values \cite{RaginskyR11,NesterovS17Random}. However, classical theoretical lower bounds \cite{Nemirovsky1984Problem} indicate that the convergence rate of many ZO methods slows down linearly with the model parameter dimension $d$, making them inefficient for high-dimensional problems. To overcome this limitation, recent research \cite{CaiMYZ22, WangDBS18,Balasubramanian18} has shown that if the gradient possesses low-dimensional structure, the query complexity can scale linearly with the intrinsic dimension and logarithmically with the number of parameters. The Memory-efficient Zero-Order Optimizer (MeZO) \cite{MalladiGNDL0A23MeZO} proposes the local effective rank assumption and theoretically proves that its per-step loss reduction does not depend on the parameter dimension $d$, but the effective Hessian rank.
\section{Overview}
\subsection{Prompt Optimization}

\begin{figure*}[t]
  \centering
  \includegraphics[width=0.95\linewidth]{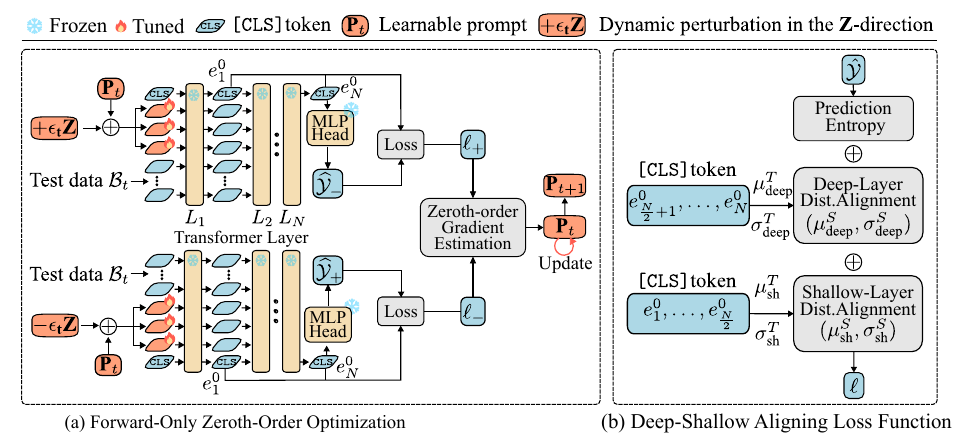}
  \caption{\textbf{Overview of our proposed FOZO.} (a) Forward-Only Zeroth-Order Optimization. This diagram illustrates how FOZO adaptively updates learnable visual prompts ($\mathbf{P}$) at test time using a zeroth-order gradient estimation. For each test batch, the model performs two forward passes by perturbing the prompt $\mathbf{P}$ in positive ($\mathbf{P} + \epsilon_t \mathbf{Z}$) and negative ($\mathbf{P} - \epsilon_t \mathbf{Z}$) directions. Here, $\mathbf{Z}$ is a perturbation vector generated from a random seed $s$, and $\epsilon_t$ is a dynamically adjusted perturbation step size. (b) Deep-Shallow Aligning Loss Function. We compute the mean ($\mu^T$) and standard deviation ($\sigma^T$) of the [CLS] token activations by grouping shallow ($e_1^0, \ldots, e_{N/2}^0$) and deep ($e_{N/2+1}^0, \ldots, e_N^0$) layers of the model, and align them with pre-computed source domain statistics ($\mu^S, \sigma^S$).
  }
  \label{fig:FOZO}
\end{figure*}

Visual-Prompt Tuning \cite{JiaTCCBHL22VPT} has demonstrated superior parameter and data efficiency in adapting pre-trained Vision Transformer (ViT) \cite{DosovitskiyB0WZ21ViT} to downstream application domains. We thus aim to leverage its efficiency and develop test-time adaptation method. 
Specifically, for an $N$-layer pre-trained ViT, the input embedding to the first layer, $\mathbf{E}_0=\{\mathbf{e}_0^j, j\in\mathbb{N},0\leq j\leq m\}$, comprises a learnable $\texttt{[CLS]}$ token $\mathbf{e}^0_0 \in \mathbb{R}^d$ and $m$ embeddings corresponding to image patches. In this setup, a set of learnable $d$-dimensional prompts $\mathbf{P} = \{\mathbf{p}^k \in \mathbb{R}^d | k \in \mathbb{N}, 1 \leq k \leq p\}$ is inserted into $\mathbf{E}_0$ before the first layer $L_1$. The subsequent forward pass through the frozen layers is defined as:
\begin{equation}
    [\mathbf{F}_1, \mathbf{E}_1] = {L_1}^\text{frozen}([\mathbf{P}^\text{learnable}, \mathbf{E}_0])
\end{equation}
\begin{equation}
    [\mathbf{F}_i, \mathbf{E}_i] = {L_i}^\text{frozen}([\mathbf{F}_{i-1}, \mathbf{E}_{i-1}])\quad i=2,3,…,N
\end{equation}
\begin{equation}
    \mathbf{\hat{y}}=\text{Head}^{\text{frozen}}(\mathbf{e}_N^0),
\end{equation}
where $\mathbf{F}_i\in\mathbb{R}^{p\times d}$ denotes the intermediate feature state of prompt $\mathbf{P}$, $\mathbf{\hat{y}}$ is the output prediction. $[\mathbf{F}_i, \mathbf{E}_i]\in\mathbb{R}^{(p+m+1)\times d}$, and $[\cdot,\cdot]$ means concatenation along the sequence length dimension. The superscripts `frozen' and `learnable' indicate the state of the parameters. 

Overall, given a batch of unlabeled test data $\mathcal{B}_t\sim\mathcal{D}_{test}$, our optimization objective is defined as:
\begin{equation}
    \min_{\mathbf{p}} \mathcal{L}(\mathbf{P}; \mathcal{B}_t),
\end{equation}
where $\mathcal{L}$ is an unsupervised learning objective.

\subsection{Forward-only Adaptation: The Challenge}
While being efficient and not altering model weights, prompt optimization usually requires backpropagation to optimize the adaptation objectives, which leads to two major challenges when developing effective test-time adaptation method:
\textit{
    1) The backward pass is computation-heavy and memory-intensive, which is not practical in many resource-limited deployment devices such as low-power FPGA chips. 
    2) Unlike prompt optimization with given set of data, data in TTA come with online and varying distributions, which may hinder accuracy and convergence of optimization.
    }

The latest SOTA backward-free prompt tuning TTA method \cite{NiuMCWZ24FOA}
formulates the adaptation process as forward-only paradigm. It optimizes a learnable $d$-dimensional prompt using Covariance Matrix Adaptation Evolution Strategy (CMA-ES), which requires only forward pass to update the input prompt.
Despite updating few learnable prompts, it is still challenging for CMA-ES algorithm to effectively optimize the prompts due to its inherent $O(d^2)$ complexity \cite{hansen2016cma}, resulting in slow convergence and suboptimal performance under the TTA data stream.
To address the above challenges and develop efficient prompt-based TTA approach, we instead focus on theoretically grounded zeroth-order gradient estimation with specifically designed optimization technique and seek to prove its convergence.

\section{Forward-Only Zeroth-Order Optimization}

In this section, we present Forward-Only Zeroth-Order Optimization (FOZO) for TTA (as illustrated in Algorithm \ref{alg:fozo_tta} and in Fig.~\ref{fig:FOZO}). We first introduce zeroth-order gradient estimation. Then, we present a convergence analysis for FOZO. Subsequently, we introduce a dynamic perturbation scheme. It achieves forward-only adaptation and addresses the optimization challenge under TTA data streams. Finally, we describe the unsupervised loss function optimized by FOZO.

\subsection{Zeroth-Order Gradient Estimation}

In the TTA scenario, unlike traditional offline training, models are updated at each time step $t$ based on the current data $\mathcal{B}_t$. We aim to estimate the gradient on the input prompts with zeroth-order estimator SPSA (Simultaneous Perturbation Stochastic Approximation)\cite{spsa}:
\begin{equation}
\begin{split}
     \widehat{g}(\mathbf{Z}) = \frac{\mathcal{L}(\mathbf{P}_t + \epsilon_t \mathbf{Z};\mathcal{B}_t) - \mathcal{L}(\mathbf{P}_t - \epsilon_t \mathbf{Z};\mathcal{B}_t)}{2\epsilon_t} \mathbf{Z},
\end{split}
\end{equation}
Where $\mathbf{Z} \sim \mathcal{N}(0, I_d) \in \mathbb{R}^d$. Nevertheless, simple one-sample of $\mathbf{Z}$ produces inaccurate estimation, we thus leverage a variant of n-SPSA gradient estimator that averages the gradient estimates over $n$ random samples of $\mathbf{Z}$:
\begin{equation}
    \widehat{\nabla} \mathcal{L}(\mathbf{P}_t;\mathcal{B}_t) = \frac{1}{n} \sum_{i=1}^{n} \widehat{g}(\mathbf{Z}_i).
\end{equation}
More accurate estimate only requires more forward passes. This is particularly suitable for resource-constrained devices.

Based on the estimated gradient, we then formulate zeroth-order optimization to update prompt as:
\begin{equation}
    \begin{split}
        \mathbf{P}_{t+1}= \mathbf{P}_t-\eta \widehat{\nabla}\mathcal{L}(\mathbf{P}_t; \mathcal{B}_t).
    \end{split}
\label{single-step update rule}
\end{equation}
Where $\eta$ is the learning rate.

\subsection{Convergence Analysis}
\label{sec:Convergence Analysis}
We then establish the convergence property of proposed zeroth-order gradient estimation based on two assumptions:

\textbf{Assumption 1 }: \textit{$\ell$-smoothness}. The loss function $\mathcal{L}$ is $\ell$-smooth, with an L-Lipschitz continuous gradient \cite{Nesterov04ConvexOptimization}. 

\textbf{Assumption 2}: \textit{Local $r$-effective rank}. \cite{MalladiGNDL0A23MeZO}. The Hessian matrix of the loss function has a low effective rank property in the local region in which the optimization process takes place. This means that, despite the high dimensionality of the model parameters $d$, the variation of the loss function is mainly concentrated in a few (r) significant directions, where $r\ll d$.

\subsubsection{Expected Loss Derivation}
Based on the $\ell$-smoothness of the loss function $\mathcal{L}$, our loss function has the standard quadratic upper bound:
\begin{equation}
\begin{split}
    \mathcal{L}(\mathbf{P}_{t+1}) & \le \mathcal{L}(\mathbf{P}_t) + \nabla\mathcal{L}(\mathbf{P}_t)^\top (\mathbf{P}_{t+1}-\mathbf{P}_t) \\
    & \quad + \frac{\ell}{2}\|\mathbf{P}_{t+1}-\mathbf{P}_t\|^2.
\end{split}
\end{equation}
Substituting the zeroth-order gradient estimation single-step update rule Eqn. \ref{single-step update rule}, into the above inequality and taking the expectation, we obtain:
\begin{equation}
\begin{split}
    \mathbb{E}_t[\mathcal{L}(\mathbf{P}_{t+1})] & \le \mathcal{L}(\mathbf{P}_t) \\
    & - \eta \mathbb{E}_t[\nabla\mathcal{L}(\mathbf{P}_t)^\top \widehat{\nabla} \mathcal{L}(\mathbf{P}_t; \mathcal{B}_t)] \\
    & \quad + \frac{\eta^2 \ell}{2}\mathbb{E}_t[\|\widehat{\nabla} \mathcal{L}(\mathbf{P}_t; \mathcal{B}_t)\|^2],
\end{split}
\end{equation}
where the expectation $\mathbb{E}_t[\cdot]$ is taken over all randomness introduced at step t, including the data batch $\mathcal{B}_t\sim\mathcal{D}_{test}$ and the SPSA random perturbation $\mathbf{Z}_t \sim \mathcal{N}(0,I_d)$.

\subsubsection{Convergence Rate} 
We first establish FOZO' expected descent with a single step and then prove its convergence over iterations.

\textbf{Theorem 1} (Expected Descent). Assuming the loss exhibits local $r$-effective rank, and referring to the relevant proofs in Supplementary B (Eqn. 15 and Eqn. 17), the expected loss can be bounded as:
\begin{equation}
\label{Convergence Analysis}
\begin{split}
    \mathbb{E}[\mathcal{L}(\mathbf{P}_{t+1})] & \leq \mathcal{L}(\mathbf{P}_t) \underbrace{- \eta \|\nabla \mathcal{L}(\mathbf{P}_t)\|^2}_{\text{True Gradient Descent}} \\
    & \quad+ \underbrace{\frac{\eta^2 \ell}{2} \gamma \mathbb{E}[\|\nabla \mathcal{L}(\mathbf{P}_t; \mathcal{B})\|^2]}_{\text{Variance Term}} \\
    & \quad+ \underbrace{C \eta \ell \epsilon_t^2 r}_{\text{Bias Term}},
\end{split}
\end{equation}

where $C$ is a constant, $\gamma = \Theta(r/n)$ is a factor related to the local $r$-effective rank \cite{MalladiGNDL0A23MeZO}, and $n$ is the number of SPSA samples. In Theorem 1, the bias term $C \eta \ell \epsilon_t^2 r$ directly quantifies the error introduced by the perturbation. For precise convergence to a stationary point, theory dictates that this bias term must eventually diminish, implying $\epsilon_t \to 0$. Conversely, in the early stages of optimization or when facing significant domain shifts, a larger $\epsilon_t$ can promote exploration, aiding navigation in noisy environments and preventing premature convergence to suboptimal local minima. We therefore propose dynamic adjustment strategy for $\epsilon_t$, decaying from $\epsilon_0$ to $\epsilon_{\text{min}}$, aiming to balance exploration and convergence in practice. The variance term represents the variance introduced by the zeroth-order estimation, whose magnitude is controlled by the effective rank $r$ rather than the parameter dimension $d$. 

\textbf{Theorem 2} (Convergence Rate): It can be proved that within $T$ iterations, the FOZO algorithm satisfies:
\begin{equation}
    \begin{split}
         \frac{1}{T} \sum_{t=0}^{T-1} \mathbb{E}[|\nabla \mathcal{L}(\mathbf{P}_t)|^2] & \leq \frac{\mathcal{L}(\mathbf{P}_0) - \mathcal{L}^*}{T\eta} \\
         & \quad + C \ell \epsilon_t^2 r + \frac{\eta \ell \gamma \sigma^2}{2}, 
    \end{split}
\end{equation}
where $\mathcal{L}^*$ is a lower bound of the loss function ($\mathcal{L}(\mathbf{P}) \ge \mathcal{L}^*$), and $\sigma^2$ is a uniform bound on the stochastic gradient variance ($\mathbb{E}_{\mathcal{B}}[\|\nabla \mathcal{L}(\mathbf{P}; \mathcal{B})\|^2] \le \sigma^2$). This indicates that FOZO's convergence rate is related to $r$, rather than parameter dimension $d$. As the number of iterations $T \to \infty$, the first term on the right-hand side $\frac{\mathcal{L}(\mathbf{P}_0) - \mathcal{L}^*}{T\eta} \to 0$.
The algorithm will eventually enter and remain within the neighborhood of a stationary point, the size of which is controlled by $\eta$ and $\epsilon_t$. We can achieve a smaller convergence neighborhood by dynamically adjusting $\epsilon_t$ and using a learning rate schedule.

\textit{For a detailed proof of Theorem 1 and 2, please refer to Supplementary B.}

\begin{table*}[t] 
  \centering
\small 
  \setlength{\tabcolsep}{0.9mm} 
  \renewcommand{\arraystretch}{0.9} 
   \begin{tabular}{@{}l*{16}{c}|lccc}  
    \toprule
     & FP & Gauss & Shot & Impul & Defcs & Gls & Mtn & Zm & Snw & Frst & Fg & Brt & Cnt & Els. & Px. & JPG & Avg & Time & Memory & \#Params\\ 
    \midrule
    NoAdapt & 1 & 57.1 & 57.0 & 57.7 & 46.8 & 35.9 & 52.8 & 45.9 & 62.4 & 62.6 & 65.8 & 78.0 & 32.1 & 45.3 & 67.0 & 67.4 & 55.57 & \textbf{94} & \textbf{819} & \textbf{0}\\
    LAME & 1 &56.7 & 56.4 & 57.4 & 46.2 & 35.2 & 52.1 & 45.1 & 58.6 & 61.8 & 62.9 & 77.7 & 24.6 & 43.9 & 66.8 & 66.9 & 54.16 & 97 & \textbf{819} & \textbf{0}\\
    T3A & 1 & 56.8 & 57.7 & 59.1 & 40.8 & 36.9 & 53.2 & 46.1 & 62.2 & 59.6 & 60.4 & 78.2 & 12.1 & 45.5 & 68.2 & 69.5 & 53.76 & 311 & 823 & \textbf{0}\\
    FOA & 2 & 57.2 & 57.5 & 58.1 & 48.6 & 37.7 & 54.3 & 47.6 & 64.2 & 65.2 & 66.8 & 78.0 & 51.3 & 48.6 & 67.7 & 69.1 & 58.13  & 224 &831 & 2304\\
    ZOA & 2 &  58.0 & 58.0 & 59.5 & 49.8 & 39.5 & 55.3 & 49.0 & 64.6 & 64.3 & 66.8 & 78.0 & 48.9 & 50.2 & 67.8 & 68.7 & 58.56  & 198 & 859 & 26145\\
    \midrule
    FOZO & 2 & 57.8 & 57.8 & 58.9 & 50.2 & 39.2 & 55.3 & 49.2 & 65.3 & 66.8 & 67.5 & 78.6 & 56.9 & 51.0 & 68.3 & 70.1 & \textbf{59.52}  &179 &831& 2304\\
    \bottomrule
    \end{tabular}
  \caption{\textit{Continual adaptation results}: Comparison with forward-only methods on ImageNet-C (5k, level 5). FP denote the number of forward propagations per single sample. For a fair comparison with the same number of forward propagations, and referencing ZOA's reproduced results, FOA's population size K is set to 2, FOZO's number of SPSA is set to 1, and ZOA's steps is set to 1. Time (s) is measured for processing ImageNet-C (5k, level 5) on an NVIDIA RTX 4090 GPU. Memory usage (MiB) is the peak GPU memory allocated by tensors, measured on a single adaptation batch. \#Params is the number of parameters updated during TTA. }
  \label{tab:continual on imagenet-c}
\end{table*}
\subsection{Dynamic Perturbation}

From our convergence analysis (Section \ref{sec:Convergence Analysis}, Theorem 1) the bias term $C \eta \ell \epsilon_t^2 r$ necessitates $\epsilon_t \to 0$ for precise convergence. However, in noisy OOD data streams or early optimization, zeroth-order gradient estimation can be inaccurate (Fig. \ref{fig:Comparison of Convergence Rates}), where a larger $\epsilon_t$ offers stronger exploration, helping escape local minima or overcome noise. Therefore, a strong condition dictates that $\epsilon_t$ must be dynamically adjusted to balance exploration and exploitation. This means employing a larger $\epsilon_t$ when gradients are large or optimization is unstable to aid exploration, and decaying $\epsilon_t$ sufficiently as the gradient norm approaches zero to ensure precise convergence.

To approximate this theoretical strong condition, we introduce a dynamic perturbation strategy for $\epsilon_t$. This strategy employs a larger $\epsilon_t$ during early adaptation or upon detecting new data domain shifts to avoid local minima, then decays $\epsilon_t$ by a fixed factor per iteration as optimization stabilizes, capped by $\epsilon_{\min}$. Additionally, $\epsilon_t$ resets to $\epsilon_0$ upon detecting new data domain shifts or optimization stalls (indicated by significant loss fluctuations). The update rule for $\epsilon$ is as follows:
\begin{equation}
\epsilon_t = \begin{cases}
\epsilon_0 & \text{if } L_t > \tau \cdot \bar{L}_t \\
\max(\epsilon_{\min}, \epsilon_{t-1} \cdot \alpha) & \text{otherwise}
\end{cases},
\label{eq:dynamically adjust}
\end{equation}
where the historical average loss $\bar{L}_t = \beta \bar{L}_{t-1} + (1 - \beta)L_t$, $\beta \in [0,1)$, and $\tau > 1$ is a predefined threshold factor.

While Theorem 2 assumes a monotonic decay $\epsilon_t \to 0$ for strict asymptotic precision, the dynamic reset in Eq.~\ref{eq:dynamically adjust} acts as a \textit{domain-aware re-initialization}. Under non-stationary TTA streams, FOZO essentially operates as a \textit{cascade of convergence processes}. Theorem 2 guarantees the performance lower-bound and convergence within each piece-wise stationary domain. Since practical TTA prioritizes rapid adaptation over asymptotic precision, this reset mechanism redirects the model for broader exploration during extreme shifts without sacrificing the optimization progress accumulated in earlier iterations.
\subsection{Loss Function}
\label{sec:loss function}
Unlike prompt learning in domain-specific data, data stream in TTA comes with no label. Thus we
design unsupervised loss function to both aligning intermediate feature statistics and minimizing prediction entropy.

\subsubsection{Aligning Deep-Shallow Feature Statistics} 
To achieve robust adaptation, we propose a group-wise Deep-Shallow alignment strategy. This involves collecting \texttt{[CLS]} token activations from both shallow ($e_1^0, \ldots, e_{N/2}^0$) and deep ($e_{N/2+1}^0, \ldots, e_N^0$) layers of the model during inference and computing their respective means and variances ($\mu^T_k, \sigma^T_k$) for each group $k \in \{\mathrm{shallow}, \mathrm{deep}\}$, and aligning them with pre-computed source domain statistics ($\mu^S_k, \sigma^S_k$). Thus the aligning objective $\mathcal{L}_{stats}$ can be expressed as:
\begin{equation}
    \mathcal{L}_{stats} = \sum_{k} \left( \| {\mu}^T_k - \mu^S_k \|_2 + \| {\sigma}^T_k - \sigma^S_k \|_2 \right).
\end{equation}

\subsubsection{Entropy Minimization Loss} 
To encourage confident predictions on the unlabeled target domain, we incorporate an entropy minimization loss, defined as:
\begin{equation}
\mathcal{L}_{ent} = - \sum_{i=1}^{B} \sum_{a=1}^{K} p_{i,a} \log(p_{i,a}),
\end{equation}
where $B$ is the batch size, $K$ is the number of classes, and $p_{i,a}$ is the predicted probability for class $a$ of sample $i$.

Together, the total loss function is defined as:
\begin{equation}
    \mathcal{L} = \lambda \mathcal{L}_{stats} + \mathcal{L}_{ent},
\end{equation}
where $\lambda$ is a hyper-parameter balancing the two objectives.
\section{Experiments}
\begin{table*}[t] 
  \centering
\small 
  \setlength{\tabcolsep}{0.75mm} 
  \renewcommand{\arraystretch}{1} 
   \begin{tabular}{@{}l*{15}{c}|lccc}  
    \toprule
      & Gauss & Shot & Impul & Defcs & Gls & Mtn & Zm & Snw & Frst & Fg & Brt & Cnt & Els. & Px. & JPG & Avg & Time & Memory & \#Params\\ 
    \midrule
    NoAdapt & 57.1 & 57.0 & 57.7 & 46.8 & 35.9 & 52.8 & 45.9 & 62.4 & 62.6 & 65.8 & 78.0 & 32.1 & 45.3 & 67.0 & 67.4 & 55.57 & 94 & 819 & 0\\
    TENT  & 57.4 & 60.1 & 60.8 & 42.0 & 41.0 & 53.7 & 49.2 & 62.7 & 59.9 & 64.1 & 78.4 & 60.0 & 48.9 & 68.4 & 68.3 & 58.32 &208 &5495& 39936\\
    DEYO  & 58.1 & 60.5 & 61.0 & 46.5 & 49.7 & 57.8 & 51.9 & 64.1 & 63.0 & 64.3 & 78.5 & 61.4 & 57.5 & 68.3 & 68.9 & 60.76 & 282 &5499& 39936\\
    SAR  & 58.3 & 60.9 & 61.3 & 48.2 & 48.7 & 58.4 & 50.7 & 62.6 & 61.9 & 64.0 & 78.4 & 59.9 & 54.6 & 68.0 & 69.4 & 60.36 &393 &5495& 29184\\
    EATA  & 57.0&61.6&61.5&52.0&48.4&59.5&52.9&63.9&61.8&66.0&78.7&62.8&54.7&69.3&69.8&61.35 &218 & 5496& 39936\\
    \midrule
    FOA (FP=28) & 57.4 & 60.8 & 62.9 & 53.1 & 45.1 & 58.3 & 53.2 & 66.5 & 67.2 & 67.6 & 80.1 & 57.6 & 51.3 & 68.5 & 70.2 & 61.32 & 2391 & \textbf{831} & \textbf{2304}\\
    ZOA (FP=28) & 58.9 & 61.6 & 63.3 & 53.8 & 46.8 & 58.9 & 53.2 & 65.2 & 61.9 & 68.3 & 78.6 & 59.2 & 56.1 & 69.5 & 71 & 61.75  & 2405 & 859& 26145\\
    FOZO (FP=26) & 58.5 & 62.1 & 63.1 & 53.0 & 48.0 & 57.4 & 55.7 & 66.1 & 68.1 & 67.3 & 79.8 & 64.5 & 54.4 & 70.1 & 70.9 & 62.60 &\textbf{2102}& \textbf{831} & \textbf{2304}\\
    FOZO (FP=28) & 58.4 & 62.8 & 64.2 & 53.1 & 43.2 & 58.3 & 56.3 & 66.5 & 68.5 & 68.3 & 80.2 & 64.1 & 54.3 & 70.3 & 71.6 & \textbf{62.67 } & 2351 &\textbf{831}& \textbf{2304}\\
    \bottomrule
    \end{tabular}
  \caption{\textit{Continual adaptation results}: Comparison with backpropagation methods on ImageNet-C (5k, level 5). When FP=28, ZOA's steps is set to 14, and FOA's population size K is 28. For FOZO, the number of SPSA $n$ is set to 13 when FP=26, and 14 when FP=28.}
  \label{tab:continual on imagenet-c with backpropagation method}
\end{table*}
\begin{table*}[t]
  \centering
\small 
  \setlength{\tabcolsep}{1.9mm} 
  \renewcommand{\arraystretch}{0.9} 
   \begin{tabular}{@{}l*{15}{c}|l} 
    \toprule
     & Gauss & Shot & Impul & Defcs & Gls & Mtn & Zm & Snw & Frst & Fg & Brt & Cnt & Els. & Px. & JPG & Avg \\ 
    \midrule
    NoAdapt  & 56.2 & 55.9 & 56.5 & 45.8 & 34.6 & 52.1 & 43.5 & 61.4 & 62.1 & 65.9 & 77.3 & 30.4 & 43.5 & 65.5 & 66.5 & 54.48 \\
    LAME & 55.8 & 55.6 & 56.2 & 45.5 & 33.6 & 51.7 & 42.7 & 58.4 & 61.1 & 64.5 & 77.1 & 16.4 & 42.0 & 64.9 & 66.3 & 52.79  \\
    T3A & 56.0 & 56.4 & 57.8 & 42.3 & 36.1 & 53.0 & 43.7 & 62.3 & 58.2 & 56.4 & 77.3 & 11.1 & 43.9 & 66.9 & 68.6  & 52.67  \\
    FOA & 56.5 & 56.5 & 57.3 & 47.7 & 36.2 & 53.1 & 45.1 & 63.2 & 64.6 & 69.9 & 77.4 & 48.6 & 46.0 & 66.2 & 67.6 & 57.07  \\
    ZOA & 57.1 & 56.8 & 57.7 & 47.8 & 37.6 & 54.1 & 46.7 & 63.4 & 62.9 & 65.3 & 77.4 & 46.2 & 46.1 & 67.0 & 67.4 & 56.91  \\
    \midrule
    FOZO &57.5 & 56.6 & 57.7 & 49.3 & 38.3 & 54.1 & 47.7 & 64.1 & 64.7 & 68.1 & 77.1 & 51.4 & 48.5 & 66.6 & 68.2 & \textbf{58.00} \\
    \bottomrule
    \end{tabular}
  \caption{Comparison with the forward-only methods on \textit{quantized} ViT models (8bit PTQ4ViT).}
  \label{tab:quant}
\end{table*}
\begin{table}[t]
    \centering
    \small
    \renewcommand{\arraystretch}{0.9} 
    \setlength{\tabcolsep}{2.4mm}
    \begin{tabular}{@{}l*{3}{ccc}}
    \toprule
     & \textbf{FP} & R  & Sketch & Memory&\#Params \\
     \midrule
     NoAdapt & 1 & 59.5  & 44.9 &\textbf{819}&\textbf{0} \\
     LAME & 1 & 58.9  & 44.4 &\textbf{819}&\textbf{0} \\
     T3A & 1 & 55.1  & 48.3 &823&\textbf{0} \\
     FOA & 2 & 63.2  & 49.7  &831&2304\\
     ZOA & 2 & 63.6 & 49.8 &859&26145\\ 
     \midrule
     FOZO & 2 & \textbf{64.1}  & \textbf{50.5 }&831&2304\\ 
    \bottomrule
    \end{tabular}
    \caption{Performance comparison on ImageNet-R and ImageNet-Sketch datasets.}
    \label{tab:reset on imagenet others}
\end{table}

\subsection{Experiment Setup}
\subsubsection{Datasets and Baselines} 
We used the following data sets to benchmark the performance of FOZO: ImageNet-C \cite{HendrycksD19imagenet-c}, ImageNetV2 \cite{RechtRSS19imagenetv2}, and ImageNet-Sketch \cite{WangGLX19imagenet-sketch}.
FOZO is primarily compared with
a set of established methods. These include forward-only approaches such as LAME \cite{BoudiafMAB22LAME}, T3A \cite{IwasawaM21T3A}, FOA \cite{NiuMCWZ24FOA}, and ZOA \cite{ZOA}. Additionally, we compare against backward-based methods like TENT \cite{WangSLOD21TENT}, EATA \cite{NiuW0CZZT22EATA}, SAR \cite{Niu00WCZT23SAR}, and DeYO \cite{lee2024DeYO}.

All experiments utilize the ViT-Base \cite{DosovitskiyB0WZ21ViT} model as the source model backbone, with its weights initialized from the timm repository \cite{rw2019timm}. For quantized models, we employ 8-bit PTQ4ViT \cite{YuanXCWS22ptq4vit}. We report the Acc@1 for all experiments. All baseline methods utilize the hyperparameters reported in their respective papers.
\subsubsection{Implementation Details} 
For fair comparison, we set the batch size of test samples to 64. We use 3 prompt embeddings, initialized uniformly. The initial learning rate $\eta$  in Eqn. 7 is 0.08. In Eqn. 12, The decay factor $\alpha$ is set to 0.9 , the threshold factor $\tau$ to 1.05, and the moving average factor $\beta$ for the historical average loss to 0.9. The trade-off parameter $\lambda$  in Eqn. 15 is set to 0.4. Following  FOA, source training statistics are estimated using the validation set of ImageNet-1K.  Please refer to Supplementary A for more implementation details.
\subsubsection{Continual Adaptation Setting}
To simulate real-world scenarios with dynamic data shifts, our main experiments are conducted under Continual Adaptation. Under this experimental setup, domain identity is unknown and no modifications are made to the model upon domain switching, i.e., continuous adaptation. We follow the RobustBench benchmark \cite{croce2021robustbench}, utilizing the ImageNet-C (5k) dataset, where only 5k images are randomly sampled per domain instead of the full 50,000. 

\subsection{Full Precision Models}


%
\subsubsection{Comparison with forward-only methods} 
Table \ref{tab:continual on imagenet-c} presents a comparison of FOZO with other forward-only methods. Under the setting of only two forward passes, our proposed FOZO, which employs prompt optimization without modifying model parameters, achieves a higher accuracy of 59.52\%. This performance surpasses FOA (58.13\%), an existing SOTA forward-only prompt optimization method, and also outperforms ZOA (58.56\%), which adapts by optimizing normalization layers. Notably, FOZO achieves this while updating only 8.8\% of the parameters compared to ZOA.

\subsubsection{Comparison with backpropagation methods} 

Table \ref{tab:continual on imagenet-c with backpropagation method}
presents a comparison of FOZO with backward-propagation algorithms, highlighting the inherent trade-off between runtime and memory consumption. Backward-propagation methods typically exhibit lower runtime but incur substantial memory overhead. Conversely, forward-only algorithms are characterized by a minimal memory footprint but often require more processing time. For a fair comparison, we align with FOA's experimental setup, employing 28 forward passes for forward-only algorithms. The experimental results demonstrate that FOZO achieves the best accuracy (62.65\%). Among forward-only methods, FOZO exhibits superior runtime performance compared to both FOA and ZOA, and furthermore, its memory footprint is lower than that of ZOA.

\subsubsection{Comparison on Additional Datasets} 
To evaluate FOZO's effectiveness across diverse datasets, we conducted experiments on two common adaptation benchmarks: ImageNet-R and ImageNet-Sketch. As shown in Tab.\ref{tab:reset on imagenet others}, FOZO maintains the best average accuracy under continuous adaptation, indicating that our dynamic perturbation strategy can effectively adapt to the distribution shifts induced by domain switching.

\subsection{Quantized Models}
In practical applications, hardware constraints, particularly on edge devices, mobile platforms, and embedded systems, impose stringent requirements on model memory footprint, computational complexity, and power consumption. Deploying full-precision (e.g., 32-bit floating-point) models in these resource-constrained environments often presents challenges such as insufficient storage, slow inference speeds, and reduced battery life. Consequently, model quantization, particularly 8-bit integer (INT8) quantization, emerges as a crucial optimization technique. By converting model weights and activations from floating-point to low-bit integers, quantization significantly reduces model size, accelerates inference, and lowers power consumption.


We compare backward-free methods on the quantized model, which does not support backward propagation well.
As shown in Table \ref{tab:quant}, both FOA and FOZO demonstrate considerable improvements over other forward-only methods. Our proposed FOZO method consistently outperforms ZOA, achieving the highest overall average Top-1 accuracy of 58.0\%, surpassing FOA's 57.07\%. 

\subsection{Ablation Studies}
\subsubsection{Component Analysis}
\begin{table}[t] 
\renewcommand{\arraystretch}{0.9} 
\centering

\label{tab:ablation_study}
\begin{tabular*}{\linewidth}{@{}lccc@{}} 
\toprule
- & Acc. (Top-1) & $\Delta$ \\
\midrule
NoAdapt & 55.1 & - \\
\midrule %
Base FOZO (ZO + Entropy) & 57.3 & +2.2 \\
+ Deep-shallow Feature Alignment & 60.1 & +2.8 \\
+ Dynamic Perturbation & 62.7 & +2.6 \\
\bottomrule
\end{tabular*}
\caption{Ablation study of FOZO components. The results are obtained with ImageNet-C (5k, level5).}
\label{tab:ablation_study}
\end{table}
Table \ref{tab:ablation_study} shows the progressive performance gains as FOZO's components are introduced. Starting from NoAdapt (55.1\%), our Base FOZO (ZO+ Entropy) improves accuracy by 2.2\%. The most substantial gain comes from Deep-Shallow Feature Statistics Alignment (2.8\%), highlighting its importance. Finally, Dynamic Perturbation (Full FOZO) provides an additional 2.6\% accuracy improvement.

\subsubsection{Convergence Rates}
To evaluate the convergence speed of our method during the adaptation, we design the following experiment: For each of the corruption in the ImageNet-C dataset, we perform a full round of model adaptation. Each adaptation round comprised 782 batches. During adaptation, we recorded the cumulative run time and calculated the average model accuracy within the batch. Subsequently, all cumulative run time and average accuracy from these 15 corruption types were averaged, the results are plotted in Figure \ref{fig:Comparison of Convergence Rates}. This clearly indicates that our proposed dynamic perturbation method can accelerate convergence speed. 


\begin{figure}[t]
    \centering
    \includegraphics[width=0.95\linewidth]{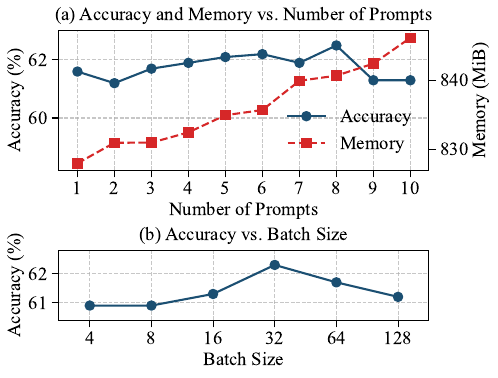}
    \caption{Ablation Study of FOZO Hyperparameters on ImageNet-C (Gaussian Noise, level 5).}
    \label{fig:hyperparameter_ablation}
\end{figure}
\subsubsection{Hyperparameters Analysis}
To evaluate FOZO's sensitivity to key hyperparameters, we conduct more ablation studies. 
Fig \ref{fig:hyperparameter_ablation}(a) shows minor accuracy variations across 1 to 10 prompts. While 8 prompts yield the highest accuracy (62.5\%), practical test-time adaptation often precludes fine-tuning prompts on unseen data. Therefore, we select 3 prompts for our main experiments, balancing performance with practical considerations. Fig \ref{fig:hyperparameter_ablation}(b) reveals the effect of batch size. Smaller batches (e.g., 4, 8, 16) show reduced accuracy (e.g., 60.9\% at batch size 4) due to higher variance in statistics and noisy zeroth-order gradient estimates. Performance peaks at batch size 32 (62.3\%), benefiting from more adaptation steps. However, increasing to 128 leads to a slight decrease (61.2\%), as large batches may over-smooth feature statistics, losing domain information. For our main experiments, we select a batch size of 64 to ensure  a fair comparison.
\section{Conclusion}
We propose Forward-Only Zeroth-Order Optimization (FOZO), a novel forward-only TTA paradigm. Compared to the latest forward-only methods, FOZO offers significant advantages: it consistently achieves higher accuracy with notably lower runtime, memory, and fewer parameters needing updates. Furthermore, both theoretical analysis and experiments demonstrate that our dynamic perturbation strategy enables faster convergence during adaptation. These advantages demonstrate that FOZO serves as a highly efficient solution, particularly suited for deployment on resource-constrained edge devices and quantized \mbox{models.}
\section*{Acknowledgment}

This work is supported by the National Science Foundation of China under Grant 62506249, the National Major Scientific Instruments and Equipments Development Project of National Natural Science Foundation of China under Grant 8 62427820, the Natural Science Foundation of Sichuan under grant 2024NSFSC1462, and the Fundamental Research Funds for the Central Universities under grant YJ202342.

{
    \small
    \bibliographystyle{ieeenat_fullname}
    \bibliography{main}
}

\clearpage
\setcounter{page}{1}
\maketitlesupplementary
\appendix 
\counterwithin{equation}{section}
\counterwithin{figure}{section}
\counterwithin{table}{section}

\section{Implementation Details}
\label{sec:Implementation Details}

\subsection{General Experiment Setup}
All experiments utilize the ViT-Base \cite{DosovitskiyB0WZ21ViT} model as the source model backbone, with its weights initialized from the timm repository \cite{rw2019timm}. For quantized models, we employ 8-bit PTQ4ViT \cite{YuanXCWS22ptq4vit}. All baseline methods utilize the hyperparameters reported in their respective papers.

\subsection{Forward-Only Adaptation Methods}
This section details the hyperparameters and specific configurations used for the back-free test-time adaptation methods evaluated.

\paragraph{FOZO} adapts the model by learning new input prompts. We use 3 prompt embeddings, initialized uniformly. We employ an n-SPSA zero-order gradient estimator, where $n$ is set to 1 as specified in Eqn. 6 for 2 forward passes, and set to 14 for 28 forward passes. The initial learning rate $\eta$  in Eqn. 7 is 0.08. In Eqn. 12, The decay factor $\alpha$ is set to 0.9 , the threshold factor $\tau$ to 1.05, and the moving average factor $\beta$ for the historical average loss to 0.9. The trade-off parameter $\lambda$  in Eqn. 15 is set to 0.4. Source training statistics are estimated using the validation set of ImageNet-1K.

\paragraph{LAME \cite{BoudiafMAB22LAME}.} We reproduce the LAME method, employing its reported hyperparameters. For Vision Transformer (ViT) features, the LAME method employs k-Nearest Neighbors (kNN) affinity, with the number of neighbors $k$ set to 5.

\paragraph{T3A \cite{IwasawaM21T3A}.} We reproduce the Test-Time Classifier Adjustment (T3A) method, adhering to its reported hyperparameters unless stated otherwise. The primary T3A hyperparameter, $M$, controls the number of low-entropy supports retained per class. While the original work explored values like $\{1, 5, 20, 50, 100, \text{N/A}\}$ (where N/A means retaining all samples), for our experiments, we consistently set $M$ to 20. 

\paragraph{FOA \cite{NiuMCWZ24FOA}.} We reproduce the FOA method, adhering to their reported hyperparameters unless stated otherwise. FOA adapts the model by learning 3 new input prompts. Optimization is performed via a derivative-free CMA-ES algorithm with a population size of 28.  The trade-off parameter $\lambda$ of fitness function is set to 0.4 for ImageNet-C, V2, and Sketch, and 0.2 for ImageNet-R. For activation shifting, the step size $\gamma$ is 1.0, and historical statistics are updated with an exponential moving average factor $\alpha = 0.1$. 

\paragraph{ZOA \cite{ZOA}.} We reproduce the ZOA method, adhering to their reported hyperparameters unless stated otherwise. For the ViT  model, we add the perturbation vectors $\epsilon$ and $\nu$with the step size  of 0.02 and 0.05, respectively. We set the learning rate of$\theta$ and $\alpha$  to be 0.0005 and 0.01, respectively. We set the maximum number of  domain knowledge parameters as N = 32. For multiple forward propagations, the steps is adjusted accordingly. For instance, 14 steps are employed for 28 forward passes.

\subsection{Back-Based Adaptation Methods}
This section details the hyperparameters and specific configurations used for the back-based test-time adaptation methods evaluated.

\paragraph{TENT \cite{WangSLOD21TENT}.} We reproduce the TENT method, adhering to their reported hyperparameters unless stated otherwise. We employ SGD with 0.9 momentum, updating only the affine parameters of layer normalization layers. The standard batch size is 64. For the ViT model, the learning rate is 0.001.
\paragraph{DeYO \cite{lee2024DeYO}.} We reproduce the DeYO method, adhering to their reported hyperparameters unless stated otherwise. We employ SGD with 0.9 momentum, updating only the affine parameters of layer normalization layers. For the ViT model, the learning rate is 0.00025. Key DeYO hyperparameters are $T_{\text{Ent}} = 0.4 \times \ln C$, $Ent_0 = 0.5 \times \ln C$, and $T_{\text{PLPD}}$ is set to 0.2.
\paragraph{SAR \cite{Niu00WCZT23SAR}.} We reproduce the SAR method, adhering to its reported hyperparameters. For the ViT model, we employ SGD with 0.9 momentum, updating only the affine parameters of Layer Normalization layers. Consistent with the original work, we freeze blocks9, blocks10, and blocks11 for adaptation. The learning rate is 0.001.
\paragraph{EATA \cite{NiuW0CZZT22EATA}.} We reproduce the EATA method, adhering to its reported hyperparameters unless stated otherwise. We employ SGD with 0.9 momentum and 0.0 weight decay, updating only the affine parameters of batch normalization layers.  The learning rate is 0.00025, and the model is updated for 1 step per batch. Key EATA hyperparameters are the entropy margin $E_0 = 0.4 \times \ln C$, the cosine similarity threshold $\epsilon = 0.05$, and the Fisher regularization trade-off parameter $\beta = 2000$. For calculating the Fisher information, 2000 source samples are utilized. The moving average factor $\alpha$ for updating model probabilities is set to 0.1.

\subsection{Continual Adaptation}
In this setup, the model undergoes continuous adaptation across all domains without re-initialization upon domain switching. This simulates a scenario where the model must continually adjust to sequential distribution shifts, building upon previous adaptations. We adhere to the RobustBench benchmark \cite{croce2021robustbench}, using the ImageNet-C (5k) dataset, where only 5,000 images are sampled per domain (instead of 50,000). This choice simulates a resource-constrained online setting and focuses on the method's ability to adapt with limited data per shift.

\subsection{Datasets}
We used the following datasets to benchmark the performance of experiments:

 \paragraph{ImageNet-C \cite{HendrycksD19imagenet-c}} is constructed by applying 15 distinct corruption types (e.g., Gaussian noise, motion blur, pixelation) across five severity levels to the original ImageNet validation set. This benchmark introduces diverse perturbations to simulate real-world image degradation, specifically designed to evaluate neural network robustness to common corruptions.
 \paragraph{ImageNet-R \cite{HendrycksBMKWDD21imagenet-r}} extends the ImageNet dataset to evaluate model generalization on non-natural or stylized images, such as cartoons, graffiti, embroidery, and sculptures. It comprises approximately 30,000 images across 200 ImageNet classes, all rendered in artistic or alternative mediums like paintings, origami, or animations.

\paragraph{ImageNet-Sketch \cite{WangGLX19imagenet-sketch}} comprises hand-drawn sketches corresponding to all ImageNet images, designed to evaluate model performance on abstract, human-drawn artistic styles.

\section{Proofs}
\label{sec:proofs}
In this section, we prove Theorems 1 and 2 proposed in the main paper. We begin by establishing the fundamental $\ell$-smoothness descent lemma. Then, we decompose the SPSA gradient estimator using Taylor expansion and analyze its bias and variance by leveraging some fundamental assumptions previously introduced. These derived bounds are then substituted back into the descent lemma, ultimately leading to the convergence analysis of FOZO. 

\subsection{Basic Theorem: $\ell$-Smoothness Descent Lemma}
\label{sec:L-Smoothness Descent Lemma}

Based on the $\ell$-smoothness assumption of the loss function $\mathcal{L}$, we have the standard quadratic upper bound:
\begin{equation}
\begin{split}
    \mathcal{L}(\mathbf{P}_{t+1}) \le \mathcal{L}(\mathbf{P}_t) & + \nabla\mathcal{L}(\mathbf{P}_t)^\top (\mathbf{P}_{t+1}-\mathbf{P}_t) \\
    & + \frac{\ell}{2}\|\mathbf{P}_{t+1}-\mathbf{P}_t\|^2,
\end{split}
\end{equation}
substituting the update rule $ \mathbf{P}_{t+1}-\mathbf{P}_t = -\eta \widehat{\nabla} \mathcal{L}(\mathbf{P}_t; \mathcal{B}_t)$, we obtain:
\begin{equation}
    \begin{split}
        \mathcal{L}(\mathbf{P}_{t+1}) \le \mathcal{L}(\mathbf{P}_t) & - \eta \nabla\mathcal{L}(\mathbf{P}_t)^\top \widehat{\nabla} \mathcal{L}(\mathbf{P}_t; \mathcal{B}_t) \\
        & + \frac{\eta^2 \ell}{2}\|\widehat{\nabla} \mathcal{L}(\mathbf{P}_t; \mathcal{B}_t)\|^2 .
    \end{split}
\end{equation}

Taking the expectation on both sides conditioned on $ \mathbf{P}_t $:
\begin{equation}
\begin{split}
\mathbb{E}_t[\mathcal{L}(\mathbf{P}_{t+1})] - \mathcal{L}(\mathbf{P}_t)&\le
- \eta \mathbb{E}_t[\nabla\mathcal{L}(\mathbf{P}_t)^\top \widehat{\nabla} \mathcal{L}(\mathbf{P}_t; \mathcal{B}_t)] \\
&\quad + \frac{\eta^2 \ell}{2}\mathbb{E}_t[\|\widehat{\nabla} \mathcal{L}(\mathbf{P}_t; \mathcal{B}_t)\|^2],
\end{split}
\label{eq:expectation terms}
\end{equation}
where the expectation $\mathbb{E}_t[\cdot]$ is taken over all randomness introduced at step $t$, including the data batch $\mathcal{B}_t \sim \mathcal{D}_{test}$ and the SPSA random perturbation $\mathbf{Z}_t \sim \mathcal{N}(0,I)$. Our task is to bound the two expectation terms on the right-hand side separately.

\subsection{Decomposing the Gradient Estimator}
To bound the expectation terms derived from the Eqn.\ref{eq:expectation terms}, we first analyze the SPSA gradient estimator. This section details its decomposition using Taylor expansion.

The SPSA gradient estimator at time $t$ is defined as:
\begin{equation}
\begin{split}
\widehat{\nabla} \mathcal{L}(\mathbf{P}_t;\mathcal{B}_t) &= \frac{1}{2\epsilon_t} \left( \mathcal{L}\left(\mathbf{P}_t + \epsilon_t \mathbf{Z}_t;\mathcal{B}_t\right) \right. \\
&\quad \left. - \mathcal{L}\left(\mathbf{P}_t - \epsilon_t \mathbf{Z}_t;\mathcal{B}_t\right) \right) \mathbf{Z}_t
\label{eq:spsa}
\end{split}
\end{equation}

A third-order Taylor expansion of  $\mathcal{L}(\mathbf{P}_t \pm \epsilon_t \mathbf{Z}_t;\mathcal{B}_t)$:
\begin{equation}
\begin{split}
\mathcal{L}(\mathbf{P}_t + \epsilon_t \mathbf{Z}_t; \mathcal{B}_t) &= \mathcal{L}(\mathbf{P}_t; \mathcal{B}_t) + \epsilon_t \mathbf{Z}_t^\top \nabla \mathcal{L}(\mathbf{P}_t; \mathcal{B}_t) \\
&\quad + \frac{\epsilon_t^2}{2} \mathbf{Z}_t^\top \nabla^2 \mathcal{L}(\mathbf{P}_t; \mathcal{B}_t) \mathbf{Z}_t \\
&\quad + \frac{\epsilon_t^3}{6} D^3\mathcal{L}(\mathbf{P}_{\xi_1}; \mathcal{B}_t)[\mathbf{Z}_t,\mathbf{Z}_t,\mathbf{Z}_t] \\
&\quad + \mathcal{O}(\epsilon_t^4),
\end{split}
\label{eq:third-order Taylor expansion1}
\end{equation}
\begin{equation}
    \begin{split}
        \mathcal{L}(\mathbf{P}_t - \epsilon_t \mathbf{Z}_t; \mathcal{B}_t) &= \mathcal{L}(\mathbf{P}_t; \mathcal{B}_t) - \epsilon_t \mathbf{Z}_t^\top \nabla \mathcal{L}(\mathbf{P}_t; \mathcal{B}_t) \\
        &\quad + \frac{\epsilon_t^2}{2} \mathbf{Z}_t^\top \nabla^2 \mathcal{L}(\mathbf{P}_t; \mathcal{B}_t) \mathbf{Z}_t \\
        &\quad - \frac{\epsilon_t^3}{6} D^3\mathcal{L}(\mathbf{P}_{\xi_2}; \mathcal{B}_t)[\mathbf{Z}_t,\mathbf{Z}_t,\mathbf{Z}_t] \\
        &\quad + \mathcal{O}(\epsilon_t^4),
    \end{split}
\label{eq:third-order Taylor expansion2}
\end{equation}
where $\mathbf{P}_{\xi_1}$ lies between $\mathbf{P}_t$ and $\mathbf{P}_t + \epsilon_t \mathbf{Z}_t$, and $\mathbf{P}_{\xi_2}$ lies between $\mathbf{P}_t$ and $\mathbf{P}_t - \epsilon_t \mathbf{Z}_t$. $D^3\mathcal{L}[\cdot,\cdot,\cdot]$ denotes the third-order derivative tensor of the loss function applied to three vectors.

Substituting Eqs. \ref{eq:third-order Taylor expansion1} and \ref{eq:third-order Taylor expansion2} into the SPSA formula, Eq. \ref{eq:spsa}, we obtain:
\begin{equation}
\begin{split}
    \widehat{\nabla} \mathcal{L}(\mathbf{P}_t; \mathcal{B}_t) &= \left( \mathbf{Z}_t^\top \nabla \mathcal{L}(\mathbf{P}_t; \mathcal{B}_t) \right) \mathbf{Z}_t \\
    &\quad + \frac{\epsilon_t^2}{12} \left( D^3\mathcal{L}(\mathbf{P}_{\xi_1}; \mathcal{B}_t)[\mathbf{Z}_t,\mathbf{Z}_t,\mathbf{Z}_t] \right. \\
    &\quad \quad \left. + D^3\mathcal{L}(\mathbf{P}_{\xi_2}; \mathcal{B}_t)[\mathbf{Z}_t,\mathbf{Z}_t,\mathbf{Z}_t] \right) \mathbf{Z}_t \\
    &\quad + \mathcal{O}(\epsilon_t^4)
\end{split}
\end{equation}

Thus, the expectation of the SPSA gradient estimate with respect to $\mathbf{Z}_t$ is:
\begin{equation}
\begin{split}
\mathbb{E}_{\mathbf{Z}_t}[\widehat{\nabla} &\mathcal{L}(\mathbf{P}_t; \mathcal{B}_t)] = \mathbb{E}_{\mathbf{Z}_t}\left[ \left( \mathbf{Z}_t^\top \nabla \mathcal{L}(\mathbf{P}_t; \mathcal{B}_t) \right) \mathbf{Z}_t \right] \\
&\quad + \frac{\epsilon_t^2}{12} \mathbb{E}_{\mathbf{Z}_t}\left[ \left( D^3\mathcal{L}(\mathbf{P}_{\xi_1}; \mathcal{B}_t)[\mathbf{Z}_t,\mathbf{Z}_t,\mathbf{Z}_t] \right. \right. \\
&\quad \quad \left. \left. + D^3\mathcal{L}(\mathbf{P}_{\xi_2}; \mathcal{B}_t)[\mathbf{Z}_t,\mathbf{Z}_t,\mathbf{Z}_t] \right) \mathbf{Z}_t \right] \\
&\quad+ \mathcal{O}(\epsilon_t^4) \\
&= \nabla \mathcal{L}(\mathbf{P}_t; \mathcal{B}_t) \\
&\quad + \frac{\epsilon_t^2}{12} \mathbb{E}_{\mathbf{Z}_t}\left[ \left( D^3\mathcal{L}(\mathbf{P}_{\xi_1}; \mathcal{B}_t)[\mathbf{Z}_t,\mathbf{Z}_t,\mathbf{Z}_t] \right. \right. \\
&\quad \quad \left. \left. + D^3\mathcal{L}(\mathbf{P}_{\xi_2}; \mathcal{B}_t)[\mathbf{Z}_t,\mathbf{Z}_t,\mathbf{Z}_t] \right) \mathbf{Z}_t \right] \\
&\quad+ \mathcal{O}(\epsilon_t^4).
\end{split}
\end{equation}

It is worth noting that the $n$-SPSA gradient estimator, $\widehat{\nabla}_{n\text{-SPSA}}$, is formed by averaging $n$ independent SPSA gradient estimators. Due to the linearity of expectation, the expected value of $\widehat{\nabla}_{n\text{-SPSA}}$ is simply the average of the expected values of these individual estimators. Since each individual SPSA estimator has the same expectation as derived above, the expected value of the $n$-SPSA estimator retains the same form. Therefore, for the sake of brevity and to avoid redundant derivations, we only present the detailed decomposition and expectation analysis for the single-sample SPSA estimator here.

\subsection{\texorpdfstring{Bounding $-\eta \mathbb{E}_t[\nabla\mathcal{L}(\mathbf{P}_t)^\top \widehat{\nabla} \mathcal{L}(\mathbf{P}_t; \mathcal{B}_t)]$}{Bounding -eta E_t[nabla L(P_t)^T hat{nabla} L(P_t; B_t)]}}
We first take the expectation of $\widehat{\nabla} \mathcal{L}(\mathbf{P}_t; \mathcal{B}_t)$ with respect to $\mathbf{Z}_t$, and then with respect to $\mathcal{B}_t$:
\begin{equation}
    \begin{split}
        \mathbb{E}_{\mathbf{Z}_t}[\widehat{\nabla} &\mathcal{L}(\mathbf{P}_t; \mathcal{B}_t)] = \mathbb{E}_{\mathcal{B}_t}\left[\mathbb{E}_{\mathbf{Z}_t}[\widehat{\nabla} \mathcal{L}(\mathbf{P}_t; \mathcal{B}_t)]\right] \\
        &= \mathbb{E}_{\mathcal{B}_t}\left[ \nabla \mathcal{L}(\mathbf{P}_t; \mathcal{B}_t) \right. \\ 
        &\quad \left. + \frac{\epsilon_t^2}{12} \mathbb{E}_{\mathbf{Z}_t}\left[ \left( D^3\mathcal{L}(\mathbf{P}_{\xi_1}; \mathcal{B}_t)[\mathbf{Z}_t,\mathbf{Z}_t,\mathbf{Z}_t] \right. \right. \right. \\ 
        &\quad \quad \left. \left. \left. + D^3\mathcal{L}(\mathbf{P}_{\xi_2}; \mathcal{B}_t)[\mathbf{Z}_t,\mathbf{Z}_t,\mathbf{Z}_t] \right) \mathbf{Z}_t \right] \right. \\ 
        &\quad \left. + \mathcal{O}(\epsilon_t^4) \right] .
    \end{split}
\end{equation}

Since $\mathbb{E}_{\mathcal{B}_t}[\nabla \mathcal{L}(\mathbf{P}_t; \mathcal{B}_t)] = \nabla \mathcal{L}(\mathbf{P}_t)$, we obtain:

\begin{equation}
\begin{split}
\mathbb{E}_t[\widehat{\nabla} &\mathcal{L}(\mathbf{P}_t; \mathcal{B}_t)] = \nabla \mathcal{L}(\mathbf{P}_t) \\
&\quad + \frac{\epsilon_t^2}{12} \mathbb{E}_t\left[ \left( D^3\mathcal{L}(\mathbf{P}_{\xi_1}; \mathcal{B}_t)[\mathbf{Z}_t,\mathbf{Z}_t,\mathbf{Z}_t] \right. \right. \\ 
&\quad \quad \left. \left. + D^3\mathcal{L}(\mathbf{P}_{\xi_2}; \mathcal{B}_t)[\mathbf{Z}_t,\mathbf{Z}_t,\mathbf{Z}_t] \right) \mathbf{Z}_t \right] \\ 
&\quad + \mathcal{O}(\epsilon_t^4)
\end{split}
\end{equation}

Let $\mathbf{b}(\mathbf{P}_t; \mathcal{B}_t)$ be the expected bias term introduced by non-zero $\epsilon_t$, defined as:
\begin{equation}
\begin{split}
\mathbf{b}(\mathbf{P}_t; \mathcal{B}_t) &= \frac{\epsilon_t^2}{12} \mathbb{E}_t\left[ \left( D^3\mathcal{L}(\mathbf{P}_{\xi_1}; \mathcal{B}_t)[\mathbf{Z}_t,\mathbf{Z}_t,\mathbf{Z}_t] \right. \right. \\
&\quad \quad \left. \left. + D^3\mathcal{L}(\mathbf{P}_{\xi_2}; \mathcal{B}_t)[\mathbf{Z}_t,\mathbf{Z}_t,\mathbf{Z}_t] \right) \mathbf{Z}_t \right] \\
&\quad + \mathcal{O}(\epsilon_t^4).
\end{split}
\end{equation}

Then the first term in the descent lemma becomes:
\begin{equation}
    \begin{split}
        -\eta \mathbb{E}_t[&\nabla\mathcal{L}(\mathbf{P}_t)^\top \widehat{\nabla} \mathcal{L}(\mathbf{P}_t; \mathcal{B}_t)]\\
        &=-\eta \nabla\mathcal{L}(\mathbf{P}_t)^\top \mathbb{E}_t[ \widehat{\nabla} \mathcal{L}(\mathbf{P}_t; \mathcal{B}_t)] \\ 
&=-\eta \nabla\mathcal{L}(\mathbf{P}_t)^\top\left(\nabla \mathcal{L}(\mathbf{P}_t)+\mathbf{b}(\mathbf{P}_t; \mathcal{B}_t) \right)\\ 
&=-\eta \|\nabla \mathcal{L}(\mathbf{P}_t)\|^2-\eta \nabla\mathcal{L}(\mathbf{P}_t)^\top \mathbf{b}(\mathbf{P}_t; \mathcal{B}_t)
    \end{split}
\end{equation}

According to Assumption 1 ($\ell$-smoothness), the third-order derivative of the loss function is bounded, typically controlled by a constant $L_H$. Simultaneously, Assumption 2 (Local Effective Rank) indicates that despite the high dimensionality $d$ of the parameter space, the curvature and higher-order variations of the loss function are primarily concentrated in a subspace with an effective dimension $r \ll d$. Therefore, $\|\mathbf{b}(\mathbf{P}_t; \mathcal{B}_t)\|$ can be bounded proportionally to $\epsilon_t^2\ell r$, where $\ell$ is the Lipschitz constant, and $L_H$ is of the same order as $\ell$. Specifically, there exists a constant $C_1$ such that $\|\mathbf{b}(\mathbf{P}_t; \mathcal{B}_t)\|\leq C_1\epsilon_t^2\ell r$.

Therefore 
\begin{equation}
    \begin{split}
        |-\eta &\nabla\mathcal{L}(\mathbf{P}_t)^\top \mathbf{b}(\mathbf{P}_t; \mathcal{B}_t)| \\
        &\leq \eta \|\nabla\mathcal{L}(\mathbf{P}_t)^\top\|\| \mathbf{b}(\mathbf{P}_t; \mathcal{B}_t)\| \\
        & \leq GC_1\epsilon_t^2\ell r =C\epsilon_t^2\ell r,
    \end{split}
\end{equation}

where $G$ is an upper bound for $\|\nabla\mathcal{L}(\mathbf{P}_t)^\top\|$, and $C=GC_1$.

Thus, the first term in the descent lemma can be bounded as
\begin{equation}
    \begin{split}
        -\eta \mathbb{E}_t&[\nabla\mathcal{L}(\mathbf{P}_t)^\top \widehat{\nabla} \mathcal{L}(\mathbf{P}_t; \mathcal{B}_t)]\\
        &\leq - \eta \|\nabla \mathcal{L}(\mathbf{P}_t)\|^2 + C\epsilon_t^2\ell r
    \end{split}
\label{eq:term1proof}
\end{equation}

\subsection{\texorpdfstring{Bounding $\frac{\eta^2 \ell}{2}\mathbb{E}_t[\|\widehat{\nabla} \mathcal{L}(\mathbf{P}_t; \mathcal{B}_t)\|^2]$}{Bounding eta^2 ell / 2 E_t[norm(hat{nabla} L(P_t, B_t))^2]}}
For the n-SPSA estimator, the expectation of its squared norm can be bounded as \cite{MalladiGNDL0A23MeZO}:
\begin{table*}[h!]
    \centering
    \small 
  \setlength{\tabcolsep}{2.1mm} 
  \renewcommand{\arraystretch}{0.9} 
    \label{tab:performance_summary}
    \begin{tabular}{@{}l*{15}{c}}
        \toprule
        Time budget & 0 & 100 & 200 & 300&400&500&600&700&800&900&1000&1100&1200&1300&1400 \\
        \midrule
         FOZO (base)& 59.5 & 59.6 & 60.0 & 60.6& 60.9&61.3&61.7&62.0&62.3&62.5&62.8&63.0&63.2&63.3 &63.4\\
         FOA&58.0&59.0&60.5&61.6&62.4&63.1&63.6&64.1&64.4&64.7&64.9&65.1&65.4&65.5&65.7 \\
         ZOA &58.7&59.7&61.2&62.5&63.4&63.8&64.1&64.3&64.6&64.8&65.0&65.1&65.2&65.3&65.4\\
         FOZO &\textbf{59.5}&\textbf{59.8}&\textbf{61.7}&\textbf{63.0}&\textbf{63.7}&\textbf{64.3}&\textbf{64.8}&\textbf{65.2}&\textbf{65.5}&\textbf{65.7}&\textbf{66.0}&\textbf{66.2}&\textbf{66.3}&\textbf{66.4}&\textbf{66.6}\\
        \bottomrule
    \end{tabular}
    \caption*{Table C.1: Detailed data on the time budget and accuracy comparison of Forward-Only Test-Time Adaptation Algorithms.}
\end{table*}
\begin{equation}
    \begin{split}
        \mathbb{E}_t[\|\widehat{\nabla} \mathcal{L}(\mathbf{P}_t; \mathcal{B}_t)\|^2] \leq \gamma \mathbb{E}[\|\nabla \mathcal{L}(\mathbf{P}_t; \mathcal{B})\|^2]
    \end{split}
\end{equation}

where $\gamma = \Theta(r/n)$ is a factor related to the effective rank. Despite the potentially large parameter dimension $d$, due to the low effective rank property of the loss function landscape, the convergence rate of zeroth-order methods does not slow down proportionally to $d$ as in classical analysis, but rather proportionally to $r$.

Therefore, the second term in the descent lemma is bounded as:
\begin{equation}
    \begin{split}
        \frac{\eta^2 \ell}{2}\mathbb{E}_t[\|&\widehat{\nabla} \mathcal{L}(\mathbf{P}_t; \mathcal{B}_t)\|^2] \\
        & \leq \frac{\eta^2 \ell}{2} \gamma \mathbb{E}[\|\nabla \mathcal{L}(\mathbf{P}_t; \mathcal{B})\|^2]
    \end{split}
\label{eq:term2proof}
\end{equation}

\subsection{Integrating into the Descent Lemma}

Substituting the bounds for the bias term and variance term back into the initial expression of the descent lemma Eqn.B.3:
\begin{equation}
    \begin{split}
        \mathbb{E}[\mathcal{L}(\mathbf{P}_{t+1})] -\mathcal{L}(\mathbf{P}_t)& \leq  - \eta \|\nabla \mathcal{L}(\mathbf{P}_t)\|^2 \\
        &+ C \eta \ell \epsilon_t^2 r \\
        & + \frac{\eta^2 \ell}{2} \gamma \mathbb{E}[\|\nabla \mathcal{L}(\mathbf{P}_t; \mathcal{B})\|^2].
    \end{split}
\end{equation}
This completes the proof of Theorem 1.
\subsection{Convergence}
The variance of the gradient noise introduced by randomly arriving mini-batches is bounded, i.e., $\mathbb{E}[\|\nabla \mathcal{L}(\mathbf{P}_t; \mathcal{B})\|^2] \le \sigma^2$. Therefore, we obtain:
\begin{equation}
    \begin{split}
        \mathbb{E}[\mathcal{L}(\mathbf{P}_{t+1})] -\mathcal{L}(\mathbf{P}_t) & \leq - \eta \|\nabla \mathcal{L}(\mathbf{P}_t)\|^2 \\
        & \quad+ C \eta \ell \epsilon_t^2 r + \frac{\eta^2 \ell \gamma \sigma^2}{2} 
    \end{split}.
\end{equation}
Summing the above inequality from $t=0$ to $T-1$, where $T$ is the number of iterations over a period of time:
\begin{equation}
    \begin{split}
        \sum_{t=0}^{T-1} \eta \mathbb{E}[\|\nabla \mathcal{L}&(\mathbf{P}_t)\|^2] \leq \sum_{t=0}^{T-1} \mathbb{E}[\mathcal{L}(\mathbf{P}_t)] \\
        &\quad - \sum_{t=0}^{T-1} \mathbb{E}[\mathcal{L}(\mathbf{P}_{t+1})] \\
        &\quad + \sum_{t=0}^{T-1} \left( C \eta \ell \epsilon_t^2 r + \frac{\eta^2 \ell \gamma \sigma^2}{2} \right) 
    \end{split}
\end{equation}
\begin{equation}
    \begin{split}
        \eta \sum_{t=0}^{T-1} \mathbb{E}[\|\nabla \mathcal{L}&(\mathbf{P}_t)\|^2] \leq \mathcal{L}(\mathbf{P}_0) - \mathbb{E}[\mathcal{L}(\mathbf{P}_T)]\\
        &+ T \left( C \eta \ell \epsilon_t^2 r + \frac{\eta^2 \ell \gamma \sigma^2}{2} \right) 
    \end{split}
\end{equation}
The loss cannot decrease indefinitely, meaning $\mathcal{L}(\mathbf{P}_T) \ge \mathcal{L}^*$. Thus, we have $\mathbb{E}[\mathcal{L}(\mathbf{P}_T)] \ge \mathcal{L}^*$:
\begin{equation}
    \begin{split}
        \eta \sum_{t=0}^{T-1} \mathbb{E}&[\|\nabla \mathcal{L}(\mathbf{P}_t)\|^2] \leq \mathcal{L}(\mathbf{P}_0) - \mathcal{L}^* \\
        &+ T \left( C \eta \ell \epsilon_t^2 r + \frac{\eta^2 \ell \gamma \sigma^2}{2} \right) 
    \end{split}
\end{equation}
Within T iterations, the upper bound for the expected average squared gradient norm is:
\begin{equation}
    \begin{split}
        \frac{1}{T} \sum_{t=0}^{T-1} \mathbb{E}[|\nabla \mathcal{L}(\mathbf{P}_t)|^2] & \leq \frac{\mathcal{L}(\mathbf{P}_0) - \mathcal{L}^*}{T\eta} \\
        &+ \underbrace{C \ell \epsilon_t^2 r  + \frac{\eta \ell \gamma \sigma^2}{2}}_{\text{Error Floor}} 
    \end{split}
\end{equation}
As the number of iterations $T \to \infty$, the first term on the right-hand side $\frac{\mathcal{L}(\mathbf{P}_0) - \mathcal{L}^*}{T\eta} \to 0$. This implies that the expected average squared gradient norm of the algorithm does not diverge, but is instead bounded by an 'Error Floor':
\begin{equation}
    \begin{split}
        \lim_{T\to\infty} \frac{1}{T} \sum_{t=0}^{T-1} \mathbb{E}[\|\nabla \mathcal{L}(\mathbf{P}_t)\|^2] \leq C \ell \epsilon_t^2 r + \frac{\eta \ell \gamma \sigma^2}{2} 
    \end{split}
\end{equation}
The algorithm will eventually enter and remain within the neighborhood of a stationary point (a point where the gradient is zero). The size of this neighborhood is controlled by $\eta$ and $\epsilon_t$.

\section{More Details on Figure 1}
\begin{table*}[h!]
    \centering
    \small 
  \setlength{\tabcolsep}{4.9mm} 
  \renewcommand{\arraystretch}{0.9} 
    \label{tab:performance_summary}
    \begin{tabular}{@{}l*{9}{c}}
        \toprule
         & $\alpha$=0.1& $\alpha$=0.2 &$\alpha$=0.3 & $\alpha$=0.4&$\alpha$=0.5&$\alpha$=0.6&$\alpha$=0.7&$\alpha$=0.8&\textbf{$\alpha$=0.9} \\
        \midrule
         ACC@1&59.13&59.15&59.04&59.34&59.33&59.45&59.47&59.50&\textbf{59.52}\\
        \bottomrule
    \end{tabular}
    \caption*{Table D.1: Effects of  the decay factor $\alpha$ in Eqn. 12. We report the continual adaptation results on ImageNet-C (5k,level 5) with 2 forward passes.}
\end{table*}
\begin{table*}[h!]
    \centering
    \small 
  \setlength{\tabcolsep}{4.6mm} 
  \renewcommand{\arraystretch}{0.6} 
    \label{tab:performance_summary}
    \begin{tabular}{@{}l*{9}{c}}
        \toprule
         & $\beta$ =0.1& $\beta$ =0.2 &$\beta$ =0.3 & $\beta$ =0.4&$\beta$ =0.5&$\beta$ =0.6&$\beta$ =0.7&$\beta$ =0.8&\textbf{$\beta$ =0.9} \\
        \midrule
         ACC@1&59.02&59.05&59.14&59.23&59.27&59.26&59.38&59.50&\textbf{59.52}\\
        \bottomrule
    \end{tabular}
    \caption*{Table D.2: Effects of the moving average factor $\beta$ in Eqn. 12. We report the continual adaptation results on ImageNet-C (5k,level 5) with 2 forward passes.}
\end{table*}
\begin{table*}[h!]
    \centering
    \small 
  \setlength{\tabcolsep}{4.6mm} 
  \renewcommand{\arraystretch}{0.6} 
    \label{tab:performance_summary}
    \begin{tabular}{@{}l*{9}{c}}
        \toprule
         & $\lambda$ =0.1& $\lambda$ =0.2 &$\lambda$ =0.3 & \textbf{$\lambda$ =0.4}&$\lambda$ =0.5&$\lambda$ =0.6&$\lambda$ =0.7&$\lambda$ =0.8&$\lambda$ =0.9 \\
        \midrule
         ACC@1&59.31&59.30&59.35&\textbf{59.52}&59.50&59.34&59.30&59.27&59.16\\
        \bottomrule
    \end{tabular}
    \caption*{Table D.3: Effects of trade-off parameter $\lambda$ in Eqn. 15. We report the continual adaptation results on ImageNet-C (5k,level 5) with 2 forward passes.}
\end{table*}
In this section, we present the detailed experimental results for each method shown in Figure 1 of the main paper.
Table C.1 present the detailed experimental results for all methods shown in Figure 1. All results were tested on ImageNet-C (level 5). To evaluate the convergence rate of different methods across all corruptions, we reset the model parameters at each domain switch. We recorded the ACC@1 for each method adapting to each corruption under the same time budget. The accuracy corresponding to the time points listed in the table is the average value across 15 corruptions under the current time budget. As evident from Table C.1 and Figure 1, FOZO achieves higher accuracy, consistently outperforming the state-of-the-art forward-only propagation methods ZOA and FOA across all different time budgets. This indicates that our dynamic perturbation method can accelerate the convergence speed, achieving higher accuracy in a shorter amount of time.
\section{More Ablation Studies}
To evaluate FOZO's sensitivity to key hyperparameters, we conducted further ablation studies. We first investigated the effect of the decay factor $\alpha$ in Eqn. 12 (see Table 1). The results indicate that as $\alpha$ decreases, the perturbation scale decays faster, which affects FOZO's exploration capability during the early stages of adaptation. Next, we analyzed the impact of the moving average factor $\beta$ in Eqn. 12 (see Table 2), finding that when $\beta$ is significantly small, the performance degrades, attributed to the generation of biased objectives. Finally, we examined the influence of the trade-off parameter $\lambda$ in Eqn. 15 (see Table 3), and ultimately selected $\lambda$=0.4 to achieve optimal performance.

\section{More Experiment: Mixed Shifts}
To further evaluate the robustness of FOZO under more diverse experimental settings, we introduce the mixed shifts scenario. This setup simulates a highly dynamic and unpredictable real-world scenario where the test data stream consists of samples from multiple, randomly interleaved domain shifts. Similar to the 'Continual Adaptation' setup, we utilize the ImageNet-C (5k) dataset for this scenario. This means that the data stream is composed of 5,000 randomly sampled images from various corruption types within ImageNet-C, presented in an unsorted, mixed fashion. This evaluates the method's robustness and agility in handling simultaneous and unannounced distribution changes with limited data per domain, reflecting a more realistic and challenging online adaptation environment.
\begin{figure}[t]
    \centering
    \includegraphics[width=1.0\linewidth]{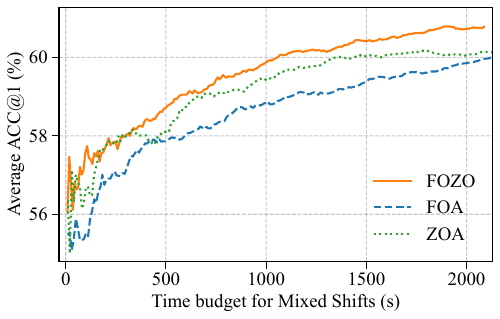}
    \caption{Mixed shift: Performance comparison on  ImageNet-C (5K, level 5).}
    \label{fig:hyperparameter_ablation}
\end{figure}

Figure E.1 compares the performance of FOZO, ZOA, and FOA on ImageNet-C (5K, level 5) under the challenging mixed shift scenario. All three methods show increasing accuracy with a larger time budget, indicating successful adaptation. Crucially, FOZO consistently outperforms the state-of-the-art forward-only prompt tuning method FOA throughout the process and surpasses ZOA's final accuracy. These results demonstrate FOZO's enhanced robustness and superior adaptation capability in highly dynamic mixed shift environments.

\end{document}